# A Hybrid Deep Learning Framework for Stock Price Prediction Considering the Investor Sentiment of Online Forum Enhanced by Popularity

Huiyu Li[a], Junhua Hu[a,*]

a: School of Business, Central South University, Changsha 410083, China

Email address: huiyu_li@csu.edu.cn (H.L.); hujunhua@csu.edu.cn (J.H.);

Corresponding author*: Junhua Hu (Email: hujunhua@csu.edu.cn)

Abstract: Stock price prediction has always been a difficult task for forecasters. Using cutting-edge deep learning techniques, stock price prediction based on investor sentiment extracted from online forums has become feasible. We propose a novel hybrid deep learning framework for predicting stock prices. The framework leverages the XLNET model to analyze the sentiment conveyed in user posts on online forums, combines these sentiments with the post popularity factor to compute daily group sentiments, and integrates this information with stock technical indicators into an improved BiLSTM-highway model for stock price prediction. Through a series of comparative experiments involving four stocks on the Chinese stock market, it is demonstrated that the hybrid framework effectively predicts stock prices. This study reveals the necessity of analyzing investors' textual views for stock price prediction.

## 1 Introduction

The stock market is an active marketplace for exchanging financial instruments and is vital to the world economy. At the heart of this market is the continuous fluctuation of stock prices, which is impacted by several factors, including corporate performance, financial indicators, and investor sentiment (Hamburger & Kochin, 1972; Shiller et al., 1984). Therefore, accurately forecasting stock prices is paramount for shareholders, financial experts, and policymakers. It enables stakeholders to make informed decisions, manage financial risks effectively, and optimize investment strategies (Kanwal et al., 2022).

Various factors such as politics, economics, markets, technology and investor behavior may cause stock prices to fluctuate, which means that the process of stock price formation is very complex (Jin et al., 2020). Because of the turbulent and erratic character of the stock market, predicting stock market movements is one of the most complex tasks in time series forecasting (Picasso et al., 2019). The efficient market hypothesis (Fama, 1991) suggests that stock price fluctuations are only affected by new information. Given the unpredictability of news, stock prices should theoretically follow a random wandering pattern, which implies that they are unpredictable. However, many studies contradict this viewpoint. These studies achieve stock price forecasting by drawing on historical experience and data. (Hafezi et al., 2015; Junqué de Fortuny et al., 2014; Lu et al., 2021; Rezaei et al., 2021; Selvin et al., 2017; Sharaf et al., 2021).

Nevertheless, accurate forecasting of stock price trends remains unresolved. Although traditional time-series methods provide valuable historical context, they cannot cover the complex elements of stock markets and the tremendous amount of information available. Therefore, they may not precisely capture the dynamics of stock price changes (Dimic et al., 2016). As a result, several researchers have attempted to improve stock price forecasting by explaining the relationship between past performance and current pricing (Guo et al., 2022; Kumar et al., 2021; Yun et al., 2021).

In recent years, the proliferation of online forum platforms such as Seeking Alpha, StockTwits, and Reddit has provided a medium for retail investors and experts to express their views on stock trends through posts and blogs. According to information cascade theory in behavioral finance (Anderson & Holt, 1997; Bikhchandani et al., 1992), individual investors tend to be influenced by their predecessors, adjusting their preferences and following their predecessors' choices. Therefore, many stock market-related opinions on online forums can shape investors' decision-making behavior and influence stock market activity. Previous studies have shown that extracting information from textual sources such as news, investor forums, and social networks can help predict stock market movements (Costola et al., 2023; Huang et al., 2023; Pedersen, 2022). That is, these texts are rich vehicles for expressing investor sentiment and opinions. Compared to numerical indicators such as survey indices and proxies, they can reflect stock market dynamics quickly and directly (Eachempati et al., 2022). Thus, a significant research question arises: how can the valuable insights from these texts be effectively utilized to predict stock prices and support investment decisions?

To this end, this study proposes a hybrid framework for stock price prediction that integrates text analysis, sentiment processing and prediction modules. The XLNET model is first fine-tuned to analyze the text of online forum posts to obtain the sentiment information of each post. Subsequently, a daily sentiment index is formed by considering the post sentiment information and the popularity factor. Finally, the sentiment index and technical indicators are merged and fed through an improved BiLSTM to predict stock prices. Our contributions can be summarized as follows:

(1) We propose a novel framework for stock price forecasting, the process of which is described above.

(2) We explore whether the introduction of the popularity factor effectively enhances investor sentiment responsiveness to stock prices.

(3) Extensive experiments on four stocks confirm the framework's efficacy and its components and improvements, such as incorporating investor sentiment for stock price prediction and utilizing the highway mechanism for capturing time-series information.

In short, this study offers new insights into stock market prediction through textual sentiment analysis, provides a comprehensive and practical approach to stock market forecasting and investment decisions, and underscores the effectiveness of retail investors' sentiment for predicting stock prices. The rest of the paper is structured as follows: Section 2 offers an overview of prior research relevant to this paper; Section 3 outlines the proposed hybrid framework for stock price prediction and provides a detailed explanation of the functioning of each module; Section 4 describes the experimental procedure; Section 5 presents the experimental results; and finally, Section 6 summarizes the findings and conclusions of this paper.

**2 Related work**

In this section, we will primarily review the research related to our paper, which mainly includes how to analyze texts on social media, time series prediction of the stock market, and the relationship between investor sentiment and the stock market.

**2.1 Text analysis on social media**

Human language is a typical form of unstructured data (Gharehchopogh & Khalifelu, 2011), and its feature extraction, analysis, and understanding are hot topics in research. The development of relevant technologies has undergone four stages: rule-based, statistical, deep learning, and large-scale pretrained language model (LPLM) (Dong, 2023).

With the widespread use of LPLM and the improvement of computer computational power, analyzing a significant amount of textual data on the internet has become possible. A typical research scenario is text sentiment analysis in the investment market. Specifically, it has been demonstrated that news affects stock price returns (Li et al., 2014), and the text analysis techniques used for this purpose have gradually shifted from dictionary-based models to LPLM, such as Transformers (Mishev et al., 2020).

Most text sentiment analysis studies focus on financial news headlines because they provide concise and critical information (Johnman et al., 2018). With the enhanced ability of the LPLM to nonlinearly map text features, some studies have started to explore the use of multisource text data, including individual investor comments, for stock price prediction (Wu et al., 2022). However, compared to news professionalism, individual investors possess typical irrational characteristics (Daniel & Titman, 1999). It is still under exploration whether the textual information they post on social media can reflect market fluctuations. Furthermore, the interpretability of social media text for the stock market is still an ongoing research topic (Gite et al., 2021).

**2.2 Time series prediction for the stock market**

Forecasting the stock market relies primarily on time series techniques and fundamental analysis. This process involves forecasting various indicators, including market fluctuations, volatility, and yield. The pertinent techniques can be classified into three main categories: regression, machine learning, and deep learning.

Earlier researchers were predominantly constrained by computer performance limitations, making it challenging for them to deal with massive amounts of data for forecasting. Consequently, they primarily relied on traditional regression models for stock market prediction. One commonly used tool is the autoregressive integrated moving average (ARIMA). For instance, Ariyo et al. (2014) employed data from two exchanges to determine the parameters of the ARIMA model, which in turn constructed a stock price prediction model. Durairaj and Krishna (2021) explored the prediction of chosen stock market values in the Philippines by employing a blend of ARIMA models, explicitly emphasizing the influence of COVID-19 on share prices. The findings show that the ARIMA model exhibits great potential for short-term forecasting, but its applicability is limited to time-series data that are either smooth or differentially smooth.

Moreover, multiple regression (MR) and logistic regression (LR) models are frequently employed in stock market prediction. Asghar et al. (2019) constructed an MR model that performs well on multiple datasets through enhanced selection of predictor variables. By optimizing the parameters of the LR model, Gong and Sun (2009) obtained better prediction results for the Shenzhen Development stock A stock than did the baseline model, including the neural network. Nonetheless, these methods heavily rely on feature selection and data quality.

Unfortunately, traditional time-series regression models exhibit two significant limitations in the stock market. Some models do not apply to datasets that deviate from the statistical assumptions, and most models perform poorly in forecasting when the input data are noisy.

To overcome the limitations of time series regression models, complex time series features have been identified using machine learning techniques to improve the accuracy of stock price forecasting. Sapankevych and Sankar (2009) introduced the support vector machine (SVM) model for stock market forecasting. Based on this, Shanthini et al. (2023) introduced a hybrid reptile search remora-based SVM approach to accurately determine stock market movements, demonstrating the importance of hybrid methods in stock market forecasting. Park et al. (2022) proposed a multitask framework that combines random forest (RF) for feature interpretation and long short-term memory (LSTM) models to predict stock market returns. Such a hybrid framework has increasingly appeared in recent studies. For example, Wang and Guo (2020) proposed combining the discrete wavelet transform with ARIMA and improving XGBoost to predict stock prices. Yun et al. (2021) constructed a hybrid framework for closing price prediction using a genetic algorithm and XGBoost. These hybrid models better address the challenges posed by high-dimensional data and improve the generalizability of machine learning for stock market time-series prediction tasks.

As an advanced machine learning method, applying deep learning techniques in time series forecasting of the stock market has recently attracted broad interest. The recurrent neural network (RNN) is a classical deep learning architecture that performs well in dealing with serial data and is widely used in stock price forecasting (Lawi et al., 2022). Moreover, the Transformer model has been gradually applied to stock price prediction (Chen et al., 2023). Furthermore, Zhang et al. (2023) found that the attention mechanism helps neural networks more accurately capture the long-term dependencies among stock market data.

**2.3 Investor sentiment and the stock market**

When applying the above techniques to stock market forecasting, previous studies have focused on technical indicators for forecasting purposes, neglecting to consider the impact of investor sentiment on market volatility. However, it is worth noting that many researchers have concluded that investor sentiment significantly affects stock market dynamics. Xu et al. (2022) found that the attitudes and emotions displayed by corporate managers in disseminating news can effectively predict the stock market, especially during periods of high emotions. Similarly, Gong et al. (2022) observed that investor sentiment generated during market volatility exhibited predictive solid power. Researchers have also made several insightful observations. For example, the emotion represented in Twitter financial tweets can considerably impact global financial indices (Valle-Cruz et al., 2022), and there is a link between music consumption activities and stock market returns (Edmans et al., 2022).

Generally, it is customary to measure investor sentiment by aggregating numerical metrics such as volatility, turnover rate, and trading volume (Gao & Martin, 2021; Lin et al., 2024; Reis & Pinho, 2021). In recent years, due to advances in the data processing capabilities of neural networks, it has become possible to introduce text and images to assess investor sentiment (Liu et al., 2023; Obaid and Pukthuanthong, 2022).

**3 Proposed framework**

**3.1 Overview**

This study introduces XSI-BiLSTM, a hybrid framework for predicting stock prices. Fig. 1 depicts the framework's three key components: the post sentiment analysis module, the sentiment index construction module, and the prediction module. First, the post-sentiment analysis module analyzes online forum post titles and bodies. Next, the sentiment index construction module generates a daily investor sentiment index, which considers the results of the above post analysis and the post popularity. This index is then combined with the stock's technical indicators and fed into the stock price prediction module to obtain predictions. The methods used in each module are described in detail below.

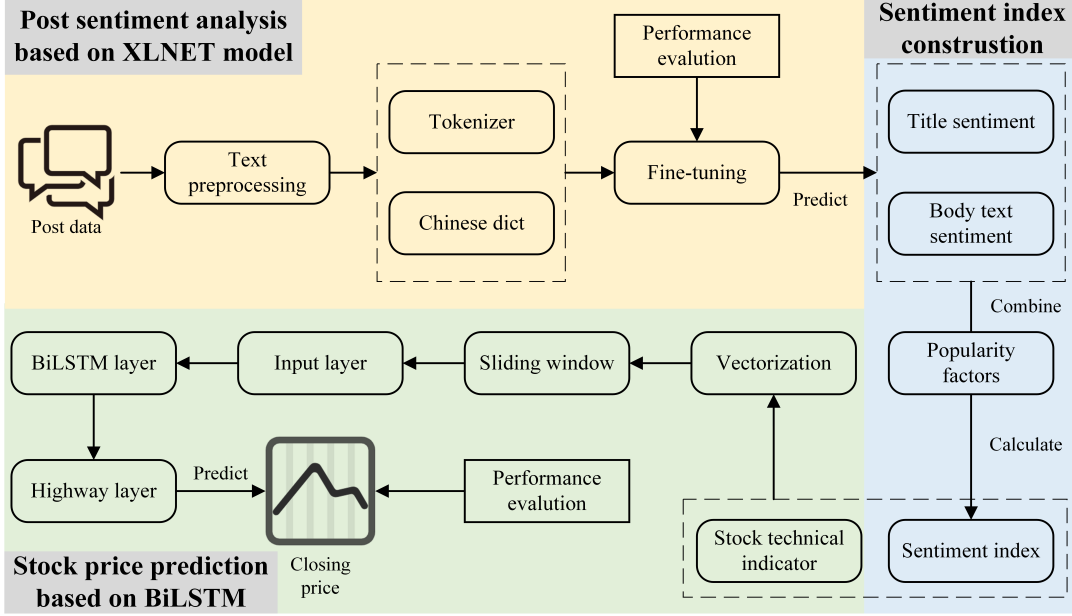

Fig. 1 The structure of XSI-BiLSTM.

**3.2 Post sentiment analysis based on the XLNET model**

XLNET is a powerful autoregressive LPLM that achieves bidirectional context learning by randomly disrupting the order of sentences (Yang et al., 2019). It randomly adopts one of the orders, masks the end words, and then predicts the masked words autoregressively. Its objective can be formulated as follows:

$$\max_{\theta} E_{z \sim Z_T} \left[ \sum_{t=1}^{T} \log p_{\theta}\left(x_{z_t} \mid x_{z_{<t}}\right) \right] \tag{1}$$

For a text sequence $x$ of length $T$, there are $T!$ permutations, where $z$ is one of all permutations $Z_T$. In addition, $t$ represents the t-th element, and $<t$ represents the first t-1 elements. For example, consider the input text "[This stock has negative news]" and its random reordering "[has, stock, negative, This, news]". The likelihood of optimization can be easily understood using Eq. (2):

$$\begin{aligned} p(z) = &\, p(has) \times p(stock \mid has) \times p(negative \mid has, stock) \\ &\times p(This \mid has, stock, negative) \times p(news \mid has, stock, negative, This) \end{aligned} \tag{2}$$

XLNET can learn to extract information about each alignment in a text sequence by sharing model parameters. However, it is impractical to compute all possible alignments due to computational constraints. Therefore, during pretraining, only one alignment is randomly selected for each input sequence. Instead of destroying sequences, XLNET uses a mask matrix to implement the alignments. This approach maintains input format consistency between the fine-tuning and downstream tasks, thus avoiding discrepancies between pretraining and fine-tuning. To compensate for the corruption caused by the mask matrix, XLNET merges the position information into the softmax function likelihood as follows:

$$p_\theta(X_{z_t} = x | x_{z<t}) = \frac{exp(e(x)^T g_\theta(x_{z<t}, z_t))}{\sum_{x'} exp(e(x')^T g_\theta(x_{z<t}, z_t))} \tag{3}$$

where $g_\theta(.)$ is the hidden state with $z_t$ location information added. Additionally, XLNET employs a two-stream self-attention mechanism to train the model effectively. This mechanism involves two streams: the Query stream, which operates with only the current location information, and the Content stream, which has access to the current content information:

$$\begin{cases} Query: g_{z_t}^m \leftarrow Attn(Q = g_{z_t}^{m-1}, KV = h_{z<t}^{m-1}; \theta) \\ Content: h_{z_t}^m \leftarrow Attn(Q = h_{z_t}^{m-1}, KV = h_{z \leq t}^{m-1}; \theta) \end{cases} \tag{4}$$

In the above attention operation, $Q$, $K$, and $V$ represent the query, key, and value, respectively. The content stream is set to the corresponding word embedding, i.e., $h_i^{(0)} = e(x_i)$. There are a total of $M = \{1, 2, \ldots, m\}$ attention layers.

For objective (1) above, only the last token in the permutation sequence is predicted during training to avoid slow convergence. Specifically, $z$ is split into a nontarget subsequence $z \leq c$ and a target subsequence $z > c$ by setting c as the cutoff point of the sequence. The objective is adjusted to:

$$\max_\theta E_{z \sim Z_T}\left[\log p_\theta(x_{z_{>c}} | x_{z_{\leq c}})\right] = E_{z \sim Z_T}\left[\sum_{t=c}^{|z|} \log p_\theta(x_{z_t} | x_{z_{<c}})\right]. \tag{5}$$

After training and fine-tuning the XLNET model, the sentiment tendency scores of the post were computed using the linear FC layer. The entire process of text prediction using XLNET is shown in Fig. 2.

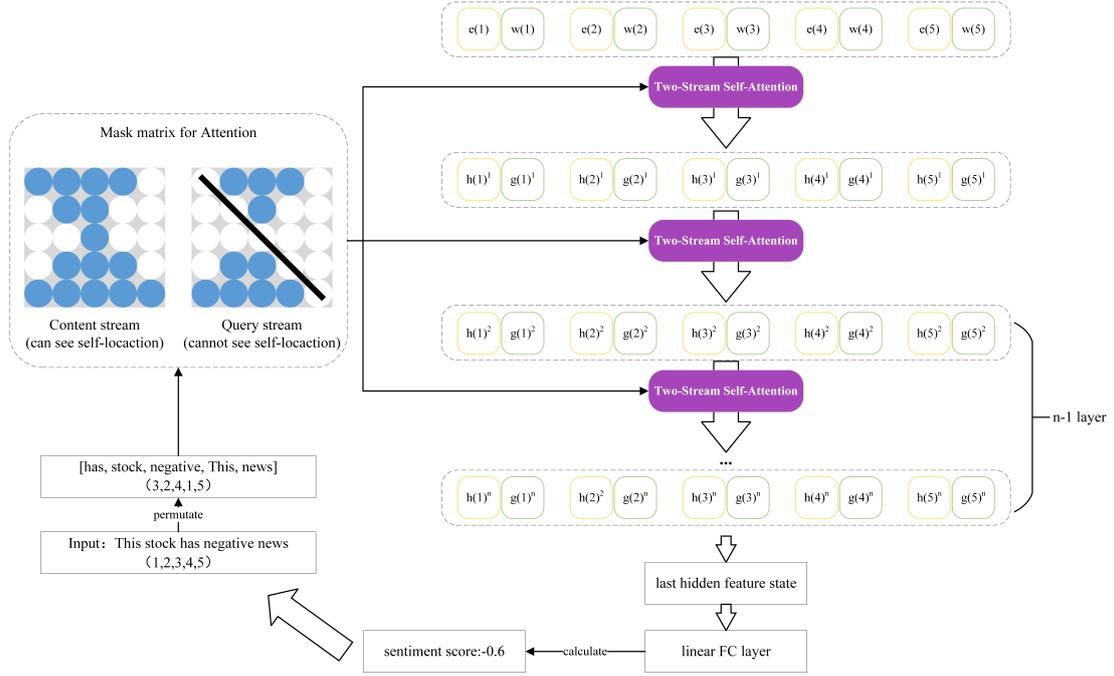

Fig. 2 XLNET-based post sentiment analysis module architecture

Because XLNET uses a more extensive Transformer-XL architecture, it can be adapted for the processing and analysis of long text sequences (Dai et al., 2019). In some natural language processing applications, such as sentiment analysis, XLNET shows promising performance, as demonstrated by our experiments. Therefore, we apply it to the text processing and analysis of posts. The specific process is outlined as follows: (1) Fine-tune XLNET using a relevant Chinese financial corpus to make it suitable for sentiment analysis of Chinese posts; (2) capture posts in web forums corresponding to the target stocks; and (3) use the fine-tuned XLNET to generate sentiment scores for the title and body of each post.

### 3.3 Construction of the sentiment index with the popularity factor

Considering the vast number of daily forum posts, feeding them directly into the predictive model would generate much noise. Therefore, it is necessary to aggregate the sentiment scores of the posts into a daily group sentiment index. Regarding the method of analyzing group sentiment, we first draw on the treatment proposed by Antweiler and Frank (2004) to compute the daily aggregated group sentiment index:

$$BI_t = \ln\left(\frac{1+M_t^{bullish}}{1+M_t^{bearish}}\right) \qquad (6)$$

where $BI_t$ represents the sentiment index for day $t$, $M^{bullish}$ is the bullish weight, and $M^{bearish}$ is the bearish weight. The original calculation method derives the weights by summing the related messages. However, this method is overly simplistic. Even when two posts share the same opinion about market fluctuations, they often differ in sentiment intensity. Therefore, it is necessary to calculate the weights in conjunction with the post sentiment tendency level. Given that the trained XLNET can provide the sentiment tendency score of the post, Eq.(6) is optimized as follows:

$$BI_t^{'} = \ln\left(\frac{1 + \sum_1^m SBull_m^t}{1 - \sum_1^n SBear_m^t}\right) \quad (7)$$

where *SBull* represents the propensity score of a post to be bullish for the stock market, with a value greater than 0, and *SBear* represents the propensity score of a post to be bearish for the stock market, with a value less than 0. Eq.(7) replaces the previous daily group sentiment calculation method by summing the propensity scores of bearish or bullish posts.

Moreover, since user behaviors, such as reading, commenting, sharing, and liking, provide insights into users' perceptions and emotional inclination toward the post (Kim & Yang, 2017), we regard them as augmentations of the post's sentiment impact. Hence, we compute the daily group sentiment using Eq. (8), taking into account the amplification effect of the post's popularity factor on its sentiment:

$$BI_t^{'} = \ln\left(\frac{1 + \sum_1^m SBull_m^t \cdot \left(R_m^{std} + C_m^{std} + L_m^{std}\right)}{1 - \sum_1^n SBear_n^t \cdot \left(R_n^{std} + C_n^{std} + L_n^{std}\right)}\right) \quad (8)$$

where $R$, $C$, and $L$ correspond to the number of reads, comments, and likes on that post, respectively. Additionally, to reduce the impact of extreme values and accelerate the training speed, standardization is used on the popularity factor with Eq. (9):

$$x_{std} = \frac{x - \bar{x}}{\sigma} \quad (9)$$

where $\bar{x}$ is the mean of the original data and $\sigma$ is the standard deviation.

To explore whether constructed sentiment indices can reflect stock price trends more accurately, we apply the Granger causality test (GCT) to examine their causal relationships. The GCT is used to determine whether one set of time series can be considered the cause of another (Granger, 1969). For time series X and Y, X is considered to be a Granger cause of Y if the addition of X's historical data in addition to Y enhances the prediction of Y, i.e., there is a Granger causality between X and Y, indicating that X helps to explain the changes in Y. The Granger causality regression equation for the effect of X on Y is shown in Eq. (10):

$$y_t = \alpha_0 + \sum_i^o \beta_i y_{t-i} + \sum_j^o \gamma_i x_{t-j} + \varepsilon \tag{10}$$

where $\varepsilon$ represents the temporal noise. In Eq. (10), if X is not a Granger cause of Y, the null hypothesis can be stated as:

$$H_0: \gamma_1 = \gamma_2 = \gamma_3 = ... = \gamma_i = 0 \tag{11}$$

In this paper, we utilize the F test to test the validity of the original hypothesis at different lag orders (denoted by $o$ in Eq. (10)), where the original hypothesis states that sentiment indices are not the cause of stock price changes. Since the post's title is exposed in the forum list and the body content is visible only after the visitors visit the post, we calculate their sentiment scores separately and name them the post title sentiment index (PTSI) and post body sentiment index (PBSI).

**3.4 Stock price prediction based on BiLSTM with a highway mechanism**

LSTM is a type of classical RNN that effectively tackles the issues of gradient explosion and vanishing gradient within the training process (Ergen & Kozat, 2018). The LSTM model comprises a memory cell ($c_t$), an input gate ($i_t$), a forgetting gate ($f_t$), and an output gate ($o_t$). Based on the hidden state at the moment $t-1$, the hidden state at the moment $t$ is computed. The steps involved in this computation can be summarized as follows:

(1) The hidden state at moment $t-1$ with the input at moment $t$ is fed to the input gate, the forgetting gate, and the output gate, respectively:

$$\begin{cases} i_t = \sigma(U_i x_t + W_i h_{t-1} + b_i) \\ f_t = \sigma(U_f x_t + W_f h_{t-1} + b_f) \\ o_t = \sigma(U_o x_t + W_o h_{t-1} + b_o) \end{cases} \tag{12}$$

where $\sigma$ is the sigmoid activation function. $U$ and $W$ are the weights to be trained, and $b$ denotes bias.

(2) Subsequently, the memory cell state was calculated:

$$c_t = f_t \odot c_{t-1} + i_t \odot \tanh(U_c x_t + W_c h_{t-1} + b_c) \tag{13}$$

where $\odot$ denotes the dot product operation and tanh is the activation function, for which Eq.(14) gives:

$$\tanh(x) = \frac{1}{1+e^{-2x}} - 1 \tag{14}$$

(3) Finally, the hidden state at moment $t$ is obtained:

$$h_t = o_t \odot \tanh(c_t) \tag{15}$$

To further enhance the capability of time-series modeling of stock prices, we employ a bidirectional LSTM (BiLSTM) to integrate the contextual information. By using two LSTM structures, BiLSTM processes both sequential and inverse order hidden states ($\vec{h}_t$ and $\overleftarrow{h}_t$) and eventually represents them in series as a long vector for further computation.

Temporal data may suffer from gradient dispersion when entering a multilayer BiLSTM network, which ultimately hinders the network's learning efficiency. We introduced a highway mechanism to allow a seamless flow of information between the front and back layers (Zilly et al., 2017). To achieve this goal, we introduce a control mechanism of carry gates in the BiLSTM network to regulate the amount of information transferred from lower memory cell states to the next level. This is realized explicitly by improving Eq.(13) as follows:

$$d_t^l = \sigma(U_d x_t^l + W_d \otimes c_t^{l-1} + b_d^l) \tag{16}$$

$$c_t^l = d_t^l \odot c_t^{l-1} + f_t^l \odot c_{t-1}^l + i_t^l \otimes \tanh(U_c^l x_t^l + W_c^l h_{t-1}^l + b_c) \tag{17}$$

In the prediction module, we employ the improved BiLSTM-highway model described above, which effectively facilitates the flow of time series information from layer to layer. Fig. 3 depicts the actual workflow of the module.

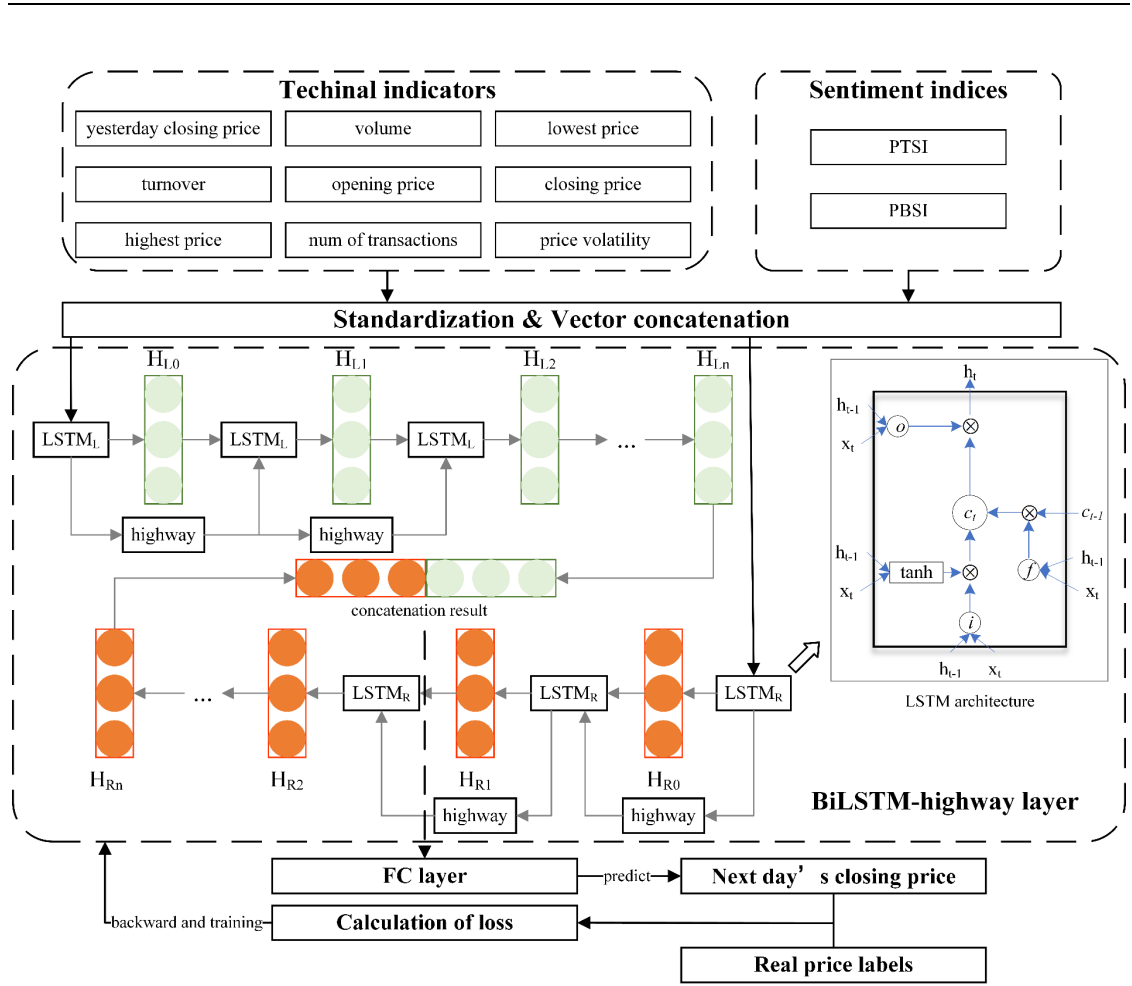

Fig. 3 The BiLSTM-highway module structure for stock closing price prediction

## 4 Experiment

### 4.1 Data sources, description and processing

#### 4.1.1 Textual data

For our proposal, two types of text data need to be collected. The first type is Chinese investor stock comments with labels, which are used to fine-tune the XLNET model. The second type is forum posts, which are used to predict the closing price of a stock in conjunction with stock technical indicators.

We collected four relevant datasets from the GitHub community[1] to form the first text data type. The researchers labeled them as positive, neutral, or negative sentiments. After summarizing the data and removing short texts, 33,329 texts were obtained. A statistical description of the data is shown in Table 1.

Table 1 Statistical description of the training data

| Category | Text quantity | Example | Translation | Label |
|---|---|---|---|---|

---

[1]. Data sources: github.com/algosenses/Stock_Market_Sentiment_Analysis, github.com/HongWooo/Sentiment, github.com/zyiyy/TextSentiment, github.com/zhy0313/double-degree.

| | | | | |
|---|---|---|---|---|
| Positive | 9554 | 没事解套很快的，耐心持有不用担心 | It is okay. You will escape the encumbrance quickly; hold on patiently and do not worry. | 1 |
| Neutral | 12975 | 建议周一再观望一下 | I suggest waiting and seeing again on Monday. | 0 |
| Negative | 10800 | 此股不建议，处于下降通道 | Do not recommend this stock. It is in a downward channel. | -1 |
| Total | 33329 | - | - | - |

We collected the second type of textual data from Eastmoney Guba[2], the largest financial online forum in China. The forum contains posts from retail and professional investors, including information such as title and body text, likes, comments and visits. We selected the following two component indices and stocks: the Growth Enterprise Index (GEI, SZ399006), the SZSE Composite Index (SZCI, SZ399001), Kweichow Moutai (KM, SH600519), and Contemporary Amperex Technology (CAT, SZ300750). These stocks are also the ones for which closing price forecasts are conducted in the experiment.

After the initial post data collection, we removed redundant emoticons, hashtags, and spaces from the post text, yielding approximately 340,000 post data points from May 31, 2022, to May 31, 2023. A statistical description of the data is given in Table 2, and an example of the post data is given in Fig. 4.

Table 2 Statistical description of the post data

| Indicator | SZ399006 | SZ399001 | SH600519 | SZ300750 | Total |
|---|---|---|---|---|---|
| Post quantity | 95392 | 55334 | 90144 | 96631 | 337501 |
| Title average length | 16.37 | 19.22 | 19.81 | 18.78 | 18.45 |
| Body text average length | 155.42 | 202.10 | 323.38 | 250.73 | 235.22 |
| The average quantity of likes | 5.17 | 7.03 | 2.73 | 3.24 | 4.27 |
| The average quantity of comments | 2.50 | 3.71 | 3.10 | 2.88 | 2.97 |
| The average quantity of views | 527.74 | 641.00 | 740.89 | 796.20 | 680.10 |

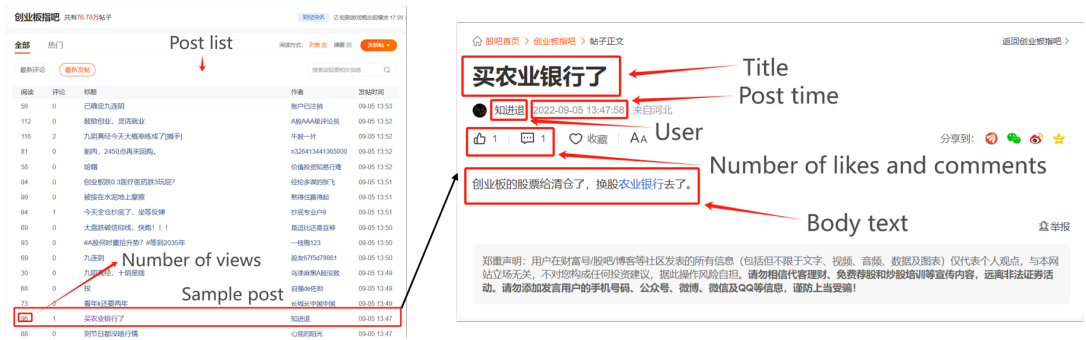

Fig. 4 Post example

### 4.1.2 Technical indicators

For the stocks introduced in Section 4.1.1, we selected nine technical indicators as their historical data for forecasting, as shown in Fig. 3. The above historical data were gathered from the RESSET database[3], and the date range is consistent with that of the textual data, which are from May 30, 2022, to June 1, 2023. Due to the stock market being closed on weekends and holidays, approximately 247 days of actual trading history data were collected. Similarly, the technical indicators are standardized according to Eq. (9).

---

[2]. Official website: guba.eastmoney.com

**4.2 Metrics**

**4.2.1 Classification evaluation metrics of the sentiment analysis model**

We evaluate the performance of the models on the sentiment classification task using accuracy, precision, recall, and F1-score as evaluation metrics. The following equations define these metrics:

$$accuracy = \frac{TP+TN}{TP+FN+FP+TN} \tag{18}$$

$$precision = \frac{TP}{TP+FP} \tag{19}$$

$$recall = \frac{TP}{TP+FN} \tag{20}$$

$$F1 = \frac{2}{\frac{1}{presicion}+\frac{1}{recall}} = \frac{2 \cdot precision \cdot recall}{precision+recall} \tag{21}$$

where *accuracy* is used to measure the ability of the model to classify samples correctly, *precision* represents the proportion of samples classified as positive cases that are positive cases, *recall* denotes how many of the samples predicted positively by the model were predicted correctly, and F1 is the harmonic mean of accuracy and recall, which may be a better metric when the categories are unbalanced.

In Eq.(18) - (21), "true positive" and "true negative" (TP and TN, respectively) relate to the ability to forecast the total number of positive or negative samples accurately. "False positive" and "false negative" (FP and FN) indicate the number of incorrectly predicted positive or negative samples, respectively.

**4.2.2 Evaluation metrics for stock price forecasting**

We employ standard regression model evaluation metrics, such as the root mean square error (RMSE), mean absolute percentage error (MAPE), and R2 score, to evaluate stock price forecasting tasks. Their expressions are shown as follows:

---

[3]. Official website: www.resset.com

$$\text{RMSE} = \sqrt{\frac{1}{n}\sum_{i=1}^{n}(y_i - \hat{y}_i)^2} \quad (22)$$

$$\text{MAPE} = \frac{100}{n}\sum_{i=1}^{n}|\frac{(y_i - \hat{y}_i)}{y_i}| \quad (23)$$

$$R2 = 1 - \frac{\sum_{i=1}^{n}(y_i - \hat{y}_i)^2}{\sum_{i=1}^{n}(\overline{y_i} - \hat{y}_i)^2} \quad (24)$$

In Eq.(22) - (24), $n$ is the total number of samples, $y_i$ and $\hat{y}_i$, respectively indicate the actual and forecasted stock price, and $\overline{y_i}$ indicates the stock's average price in the testing set.

**4.3 Experiment settings**

To ensure the superiority of the model we proposed, support vector regression (SVR), BiLSTM, LSTM-Att, GRU-Att, and Transformer are selected as benchmark models (Smola and Schölkopf 2004; Althelaya et al. 2018; Zhang et al. 2019; Niu and Xu 2020; Vaswani et al. 2017). In addition, to ensure that XLNET can precisely analyze investor sentiment from text, we use SVM, naive Bayes classification (NB), TextCNN, and BERT as comparison models (Joachims, 1998; Flach & Lachiche, 2004; Yuan et al., 2021; Devlin et al., 2018). The hyperparameter settings of XSI-BiLSTM are provided in Table 3.

Table 3 Hyperparameter settings of XSI-BiLSTM

| Module | Hyperparameter | Parameter |
|---|---|---|
| Post-sentiment analysis module (Based XLNET) | Batch size | 16 |
| | Max length | 128 |
| | Epochs | 100 with earlystop |
| | Warm up ratio | 0.1 |
| | Learning rate | 2e-5 |
| | Dropout | 0.2 |
| Prediction module (Based BiLSTM) | Batch size | 32 |
| | Hidden layer size | 128 |
| | Learning rate | 1e-3 |
| | Weight decay | 1e-2 |
| | Epochs | 1000 with earlystop |

**5 Results**

**5.1 Training and comparison of sentiment analysis models**

We divide the first type of text data into a training set, a testing set and a validation set at a ratio of 8:1:1 to compare the performance of each model horizontally. Cross-entropy is used as the loss function during the training process, as shown in Eq. (25). Table 4 shows the results of each model.

$$L = -\sum_{i=1}^{m} p(x_i) \log(q(x_i)) \tag{25}$$

where $p(x_i)$ denotes the actual label with a value of 1 or 0. $q(x_i)$ is the probability of belonging to the category as output by the model after softmax computation, and $m$ is the number of samples.

Table 4 Performance results of five sentiment analysis models. The best results are highlighted in bold, followed by the same.

| Metrics | XLNET | SVM | NB | Bert | TextCNN |
|---|---|---|---|---|---|
| Accuracy | **81.64** | 44.97 | 74.20 | 81.34 | 76.12 |
| Precision | **82.73** | 45.07 | 77.20 | 82.33 | 77.06 |
| Recall | **81.49** | 44.98 | 72.79 | 81.24 | 75.47 |
| F1-Score | **81.97** | 44.52 | 73.43 | 81.68 | 75.99 |

As shown in Table 4, XLNET performed the best in the sentiment analysis task. The F1 score of the XLNET, which represents comprehensive performance, is 81.97. The performance of BERT is similar to that of XLNET, with an F1 score of 81.68. Generally, the LPLM is superior to ordinary neural network models or traditional machine learning techniques (Han et al., 2021), which is also reflected in our comparison results. In addition, the LPLM has a breakneck convergence speed in transfer learning. Fig. 5 shows the changes in training and testing loss of XLNET with the number of epochs. A good fit was achieved after only a dozen epochs.

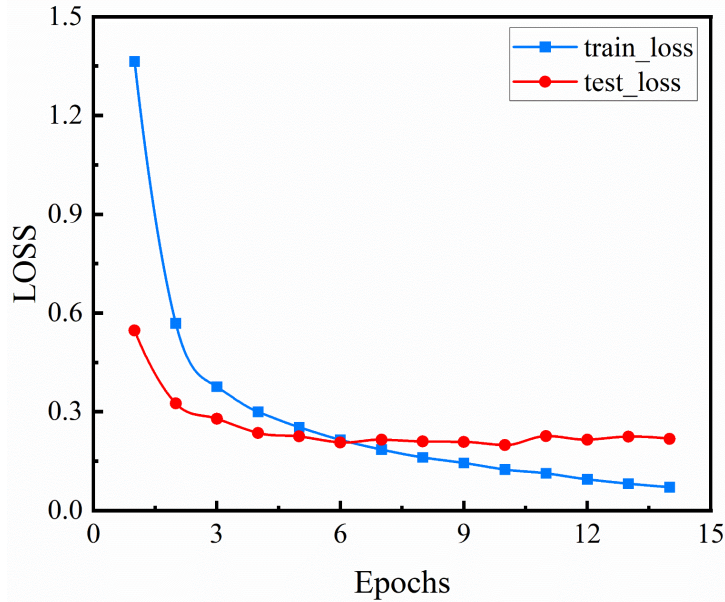

Fig. 5 Changes in loss during the training process of the XLNET model over time

The comparison results show that applying XLNET to analyze the hidden sentiment in textual data is effective, and its performance is the same as that of BERT. Considering that the structure of the XLNET is more suitable for processing long text (Sweidan et al., 2021), it is reasonable to believe that the XLNET can better analyze the sentiment of a post, which often contains hundreds of words in the body of the post, compared to other models.

Then, we use the fine-tuned XLNET to analyze the post sentiment and output scores through the final FC linear layer, with a range of $[-1,1]$. When the score is closer to -1, 0, or 1, the post's sentiment toward stock prices tends to be bearish, neutral, or bullish, respectively.

**5.2 Sentiment validity verification**

In this section, we present the results of the GCT to showcase the effectiveness of utilizing the sentiment index in predicting stock closing prices. Additionally, we analyze whether sentiment indices better capture stock market fluctuations when considering the popularity factor. It is essential to confirm the stability of the time series before performing GCT. To accomplish this, we apply the augmented Dickey-Fuller (ADF) test to verify the time series, and the results are displayed in Table 5. Generally, the stock closing price exhibits a nonsmooth series. Hence, we employ the price rate of change (ROC) as the test variable.

Table 5 ADF test results

| Variables | SZ399006 | SZ399001 | SH600519 | SZ300750 |
|---|---|---|---|---|
| ROC | -10.146*** | -10.17*** | -6.319*** | -10.265*** |
| PTSI | -3.065** | -6.532*** | -6.125*** | -4.616*** |
| PBSI | -2.903** | -6.203*** | -6.129*** | -4.779*** |
| PTSI' | -4.314** | -10.558*** | -5.93*** | -8.681*** |
| PBSI' | -2.807* | -8.302*** | -7.086*** | -5.48*** |

Note: ***, **, and * represent significance at the 1%, 5%, and 10% levels respectively, followed by the same.

According to Table 5, the ROC, PTSI, PBSI, PTSI', and PBSI' of the four stocks exhibit ADF significance, making them suitable for GCTs. Among these indices, the PTSI and PBSI are sentiment indices that do not consider the popularity factor, whereas the PTSI' and PBSI' are considered. The GCTs with different lag durations are shown in Table 6.

Table 6 Results of GCTs with lags of 1, 2, and 3 days

| Stock | Variables | Sentiment index does not | ROC does not Granger cause |
|---|---|---|---|

| code | | Granger cause ROC | | | sentiment index | | |
|---|---|---|---|---|---|---|---|
| | | lags(1) | lags(2) | lags(3) | lags(1) | lags(2) | lags(3) |
| SZ399006 | PTSI | 0.79 | 1.13 | 0.48 | 127.64*** | 54.79*** | 25.51*** |
| | PBSI | 0.48 | 2.56* | 0.94 | 180.17*** | 79.59*** | 31.91*** |
| | PTSI' | 6.01** | 3.38*** | 0.62 | 112.25*** | 48.24*** | 22.61*** |
| | PBSI' | 0.82 | 2.58* | 0.75 | 172.73*** | 80.69*** | 34.56*** |
| SZ399001 | PTSI | 0.33 | 0.33 | 0.04 | 132.69*** | 56.97*** | 23.66*** |
| | PBSI | 0.19 | 0.06 | 0.08 | 172.41*** | 67.17*** | 25.52*** |
| | PTSI' | 2.77* | 2.85* | 1.22 | 96.32*** | 79.81*** | 46.77*** |
| | PBSI' | 1.56 | 0.43 | 0.54 | 122.75*** | 87.27*** | 48.87*** |
| SZ600519 | PTSI | 0.09 | 1.85 | 0.17 | 103.41*** | 39.20*** | 15.89*** |
| | PBSI | 0.65 | 1.98 | 0.92 | 134.45*** | 57.27*** | 24.06*** |
| | PTSI' | 3.67* | 1.16 | 1.62 | 81.83*** | 62.57*** | 30.93*** |
| | PBSI' | 0.08 | 0.60 | 0.81 | 99.15*** | 91.85*** | 41.83*** |
| SZ300750 | PTSI | 1.38 | 0.17 | 0.82 | 222.91*** | 108.06*** | 47.50*** |
| | PBSI | 1.42 | 0.48 | 1.67 | 324.65*** | 157.37*** | 79.07*** |
| | PTSI' | 2.14 | 1.61 | 2.13* | 87.73*** | 87.25*** | 24.52*** |
| | PBSI' | 3.04* | 0.69 | 0.60 | 227.66*** | 228.34*** | 56.50*** |

According to Table 6, stock price fluctuation is a significant factor causing a change in users' posting sentiment in online forums, and all the indices are significant at the 1% level at different lag orders. In other words, the sentiment index hidden under the post text can effectively reflect users' reactions to stock prices. Interestingly, the PBSI shows more robust Granger causality at one lag, while the PBSI' is more significant at 2 and 3 lags. This may imply that the former is more effective in reflecting users' perceptions of stock prices during short periods of stock price changes. However, in the long run, the essential posts that gain more recognition are more representative of popular tendencies.

In contrast, the PTSI and PBSI have fewer Granger causality effects on stock price changes. Only in the lagged 2-period instance of SZ399006 does the PBSI show significance at the 10% level. However, after considering the popularity factor, both PTSI' and PBSI' are significant at specific lag periods for the four stocks. One possible explanation is that the diversity of information in a forum may interfere with judgment if it is not selectively received. When the popularity factor is incorporated, it allows posts that receive more attention and recognition to carry more weight in sentiment index calculations, reflecting more realistic investor sentiment. These sentiments are further reflected in the public's subsequent decision-making behavior in the stock market, which impacts stock prices. This finding is supported by the fact that PTSI' has a more significant impact than does PBSI'. In online forums, post titles appear in a list. Generally, users click and participate in post discussions only if they are interested in the ideas in the title, thus increasing the popularity of the post. That is, post titles will be more influenced by popularity. The significance of the coefficients suggests that PTSI' is more reflective of most investors' inclination toward future stock prices than is PBSI', which verifies this view.

In conclusion, investor sentiment effectively reflects investor perceptions of changes in closing prices. After incorporating the popularity factor, bidirectional Granger causality with stock prices is established. Therefore, its application in predicting stock closing prices is justified.

**5.3 Prediction results and comparison**

We divide the stock data into training and testing sets at a ratio of 8:2 and use the historical data from 7, 15, and 30 days before the target prediction date as the inputs to the model. The predicted prices of the models are compared with the actual stock prices using the introduced metrics to evaluate the effectiveness of the XSI-BiLSTM framework proposed in this paper.

Table 7 Detailed metrics comparison of each stock under six models

| Input windows | | 7 | | | 15 | | | 30 | | |
|---|---|---|---|---|---|---|---|---|---|---|
| Stock | Model | RMSE | MAPE | R2 | RMSE | MAPE | R2 | RMSE | MAPE | R2 |
| SZ399006 | SVR | 35.91 | 1.27% | **0.7776** | 48.83 | 1.67% | 0.5889 | 60.93 | 2.13% | 0.3600 |
| | GRU-Att | 55.84 | 1.96% | 0.4624 | 43.13 | 1.40% | 0.6792 | 57.01 | 2.08% | 0.4397 |
| | BiLSTM | 47.20 | 1.68% | 0.6158 | 43.55 | 1.53% | 0.6730 | 51.98 | 1.83% | 0.5341 |
| | LSTM-Att | 56.84 | 2.02% | 0.4430 | 49.27 | 1.71% | 0.5815 | 50.11 | 1.77% | 0.5670 |
| | Transformer | 66.21 | 2.48% | 0.2443 | 64.75 | 2.49% | 0.2772 | 74.25 | 2.72% | 0.0496 |
| | XSI-BiLSTM | **35.06** | **1.23%** | 0.7773 | **36.65** | **1.27%** | **0.7566** | **40.32** | **1.52%** | **0.7055** |
| SZ399001 | SVR | 198.91 | 1.39% | 0.6914 | 242.67 | 1.71% | 0.5406 | 270.79 | 2.10% | 0.4280 |
| | GRU-Att | 187.39 | 1.42% | 0.7261 | 191.02 | 1.44% | 0.7154 | 136.16 | 0.99% | 0.8554 |
| | BiLSTM | 124.67 | 0.83% | 0.8788 | 123.85 | 0.84% | 0.8804 | 112.83 | 0.76% | 0.9007 |
| | LSTM-Att | 174.42 | 1.31% | 0.7627 | 121.79 | 0.85% | 0.8843 | 216.34 | 1.51% | 0.6349 |
| | Transformer | 187.65 | 1.35% | 0.7253 | 160.97 | 1.17% | 0.7979 | 159.55 | 1.15% | 0.8014 |
| | XSI-BiLSTM | **100.60** | **0.67%** | **0.9173** | **99.92** | **0.71%** | **0.9184** | **99.84** | **0.68%** | **0.9186** |
| SH600519 | SVR | 23.42 | 1.09% | 0.6966 | 28.04 | 1.25% | 0.5653 | 31.16 | 1.41% | 0.4630 |
| | GRU-Att | 29.30 | 1.45% | 0.5254 | 24.93 | 1.11% | 0.6563 | 30.15 | 1.42% | 0.4972 |
| | BiLSTM | 23.36 | 1.10% | 0.6981 | 23.97 | 1.03% | 0.6822 | 23.12 | 1.09% | 0.7043 |
| | LSTM-Att | 24.13 | 1.06% | 0.6779 | 30.59 | 1.43% | 0.4825 | 34.82 | 1.67% | 0.3293 |
| | Transformer | 27.76 | 1.22% | 0.5738 | 37.55 | 1.88% | 0.2201 | 33.71 | 1.55% | 0.3714 |
| | XSI-BiLSTM | **20.76** | **0.91%** | **0.7297** | **20.87** | **0.94%** | **0.7269** | **20.29** | **0.89%** | **0.7418** |
| SZ300750 | SVR | 6.43 | 2.33% | 0.1835 | 6.56 | 2.33% | 0.1507 | 6.24 | 2.29% | 0.2301 |
| | GRU-Att | 4.41 | 1.58% | 0.6166 | 4.71 | 1.74% | 0.5615 | 5.51 | 1.96% | 0.3994 |
| | BiLSTM | 4.57 | 1.66% | 0.5868 | 4.62 | 1.67% | 0.5782 | 4.44 | 1.62% | 0.6110 |
| | LSTM-Att | 4.65 | 1.61% | 0.5722 | **4.02** | **1.49%** | **0.6804** | 5.01 | 1.73% | 0.5039 |
| | Transformer | 5.88 | 2.00% | 0.3167 | 6.01 | 2.24% | 0.2871 | **4.47** | **1.57%** | 0.6055 |
| | XSI-BiLSTM | **3.94** | **1.43%** | **0.7172** | 4.18 | 1.52% | 0.6809 | 4.40 | 1.58% | **0.6467** |

Table 7 gives a quantitative assessment of the performance of the different models. Among them, XSI-BiLSTM achieves the best performance in most cases. For example, for the MAPE, which can most intuitively reflect the difference between forecasted and actual stock prices, the XSI-BiLSTM outperforms the other five baseline models by 67.03%, 46.06%, 17.73%, 44.08%, and 67.24%, respectively. This preliminary result confirms the effectiveness of our proposed framework. Notice that XSI-BiLSTM does not have a significant performance lead in SZ300750, which may be because CAT is characterized by a more pronounced cyclicality than the other stocks (Fig. 6d illustrates this cyclical variation). This cyclical volatility is likely to be successfully captured by the attention mechanism. Therefore, LSTM-Att and Transformer achieve good prediction performance.

Interestingly, as the input window time increases, the predictive performance of the models does not significantly improve, and SVR even shows a decrease in performance. This may be because stock prices are nonstationary, and the SVR model is not complex enough to accommodate a more extended time series. However, XSI-BiLSTM exhibits superior predictive performance under either short-term or long-term input windows, which is attributed to the timely feedback and adjustment of investor sentiment.

The prediction results of each model on the testing set (March 21, 2023 to May 31, 2023) are shown in Fig. 6. The results show that XSI-BiLSTM maintains good tracking performance in predicting the closing prices of the four stocks without significant bias. SVR performs well in the early period but shows serious bias later. This phenomenon may be because the features learned by SVR during the training process are only applicable to stock price movements in the time frame covered by the training set. Consequently, they cannot accurately capture the changes in stock price dynamics on the date the testing set is located. This leads to a bias between the prediction results and the actual stock prices. The GRU-Att and LSTM-Att models exhibit strong performance with respect to specific stocks. However, they demonstrate significant bias in others, likely stemming from a misallocation of weights within the attention mechanism. Transformer performs well in most cases but does not control the magnitude of the change when there is a change in the stock price trend. This may be caused by its self-attention mechanism focusing too much on short-term fluctuations. BiLSTM's tracking ability is superior. Nevertheless, it captures changes in conditions more slowly than XSI-BiLSTM, possibly due to its lack of feedback on market sentiment, resulting in slower adjustment.

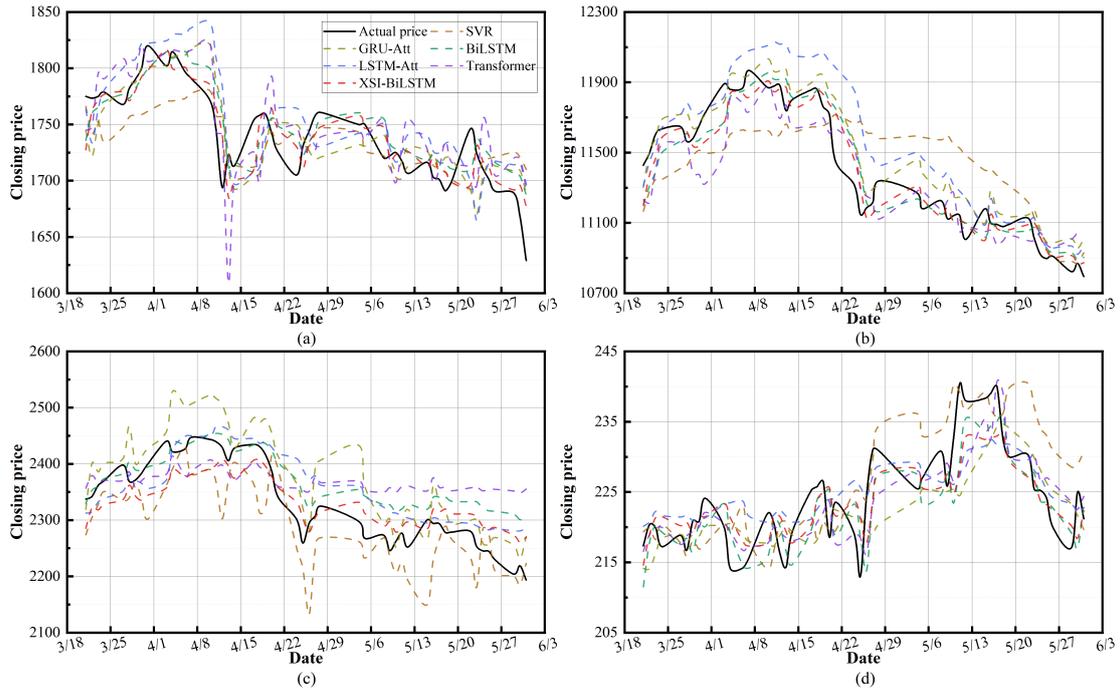

Fig. 6 Actual closing price and predicted price of four stocks under different models.
(a): SZ399006 (b): SZ399001 (c): SH600519 (d): SZ300750

Fig. 7 - 10 show each model's relative percentage error (RPE) and average one-sided error (AOSE) between the predicted and actual closing prices. Comparatively, BiLSTM, Transformer, and XSI-BiLSTM exhibit relatively small and stable overall biases. However, the RPE of the Transformer is more severe on certain trading days, e.g., April 25, 2023 for SZ399006; April 13, 2023 for SZ600519; and May 11, 2023 for SZ300750. In some cases, the GRU-Att and LSTM-Att models are too "optimistic" and predict stock prices much higher than their actual values. Upon reversal of the stock price, the RPE of most models becomes more significant.

In contrast, the XSI-BiLSTM model, which integrates investor sentiment, exhibits superior performance. This enhancement may be because people's judgment of future stock prices is reflected in textual sentiment, and accurate bias correction can be achieved by using textual sentiment, which allows the XSI-BiLSTM to achieve a more balanced AOSE. In short, the stock price prediction task is highly complex, and predicting the price based on the time-series characteristics of the stock's technical indicators alone may not be accurate enough.

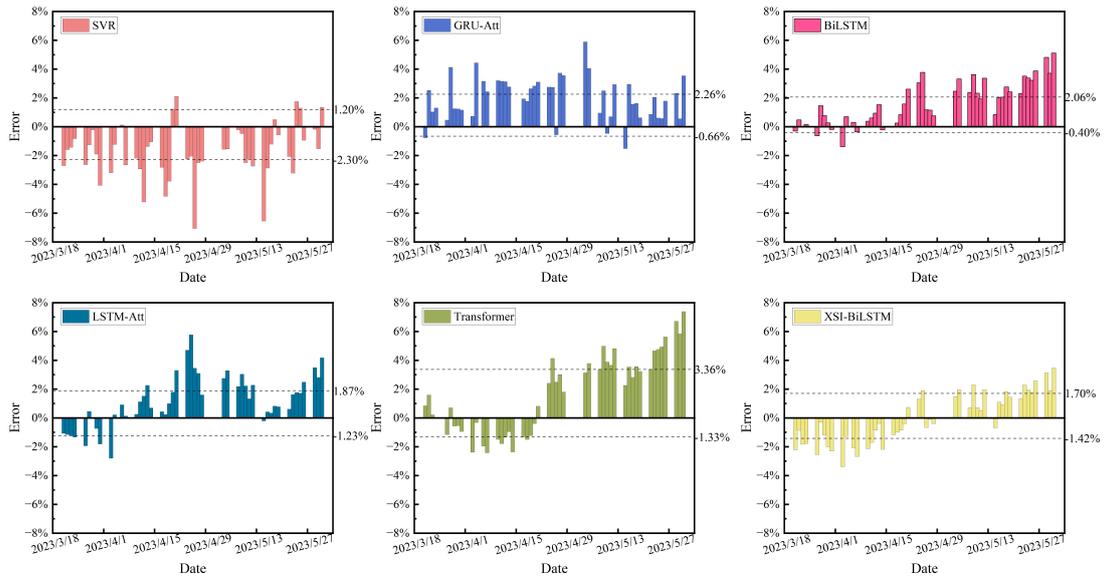

Fig. 7 RPE of each model for SZ399006

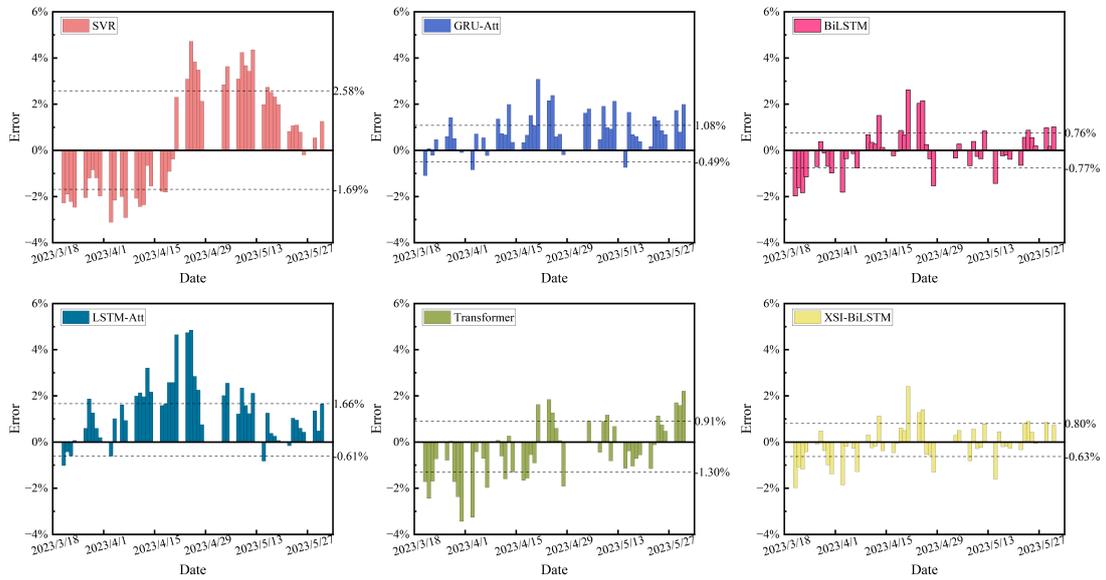

Fig. 8 RPE of each model for SZ399001

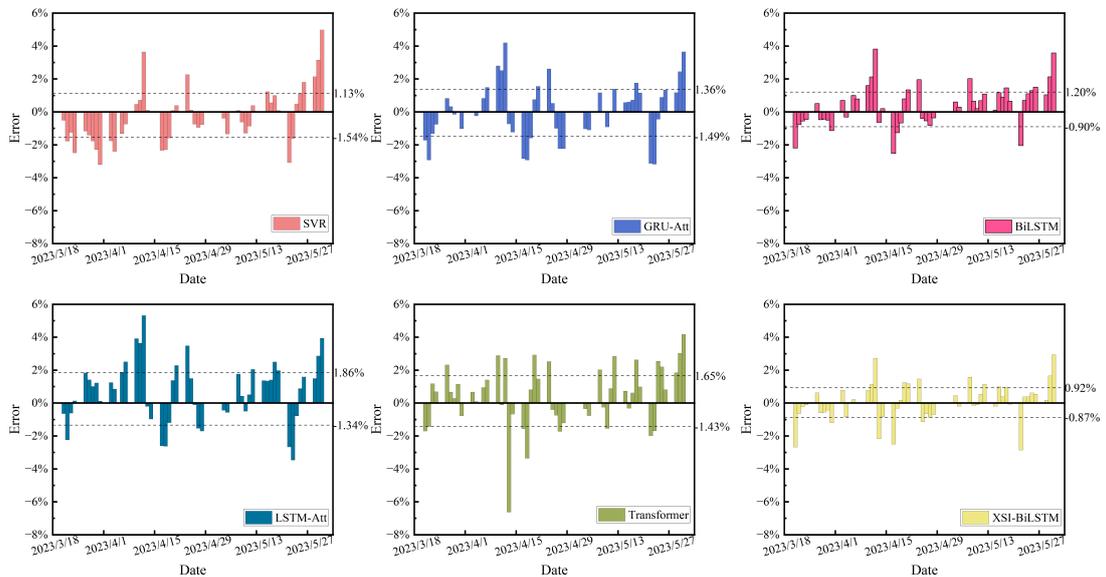

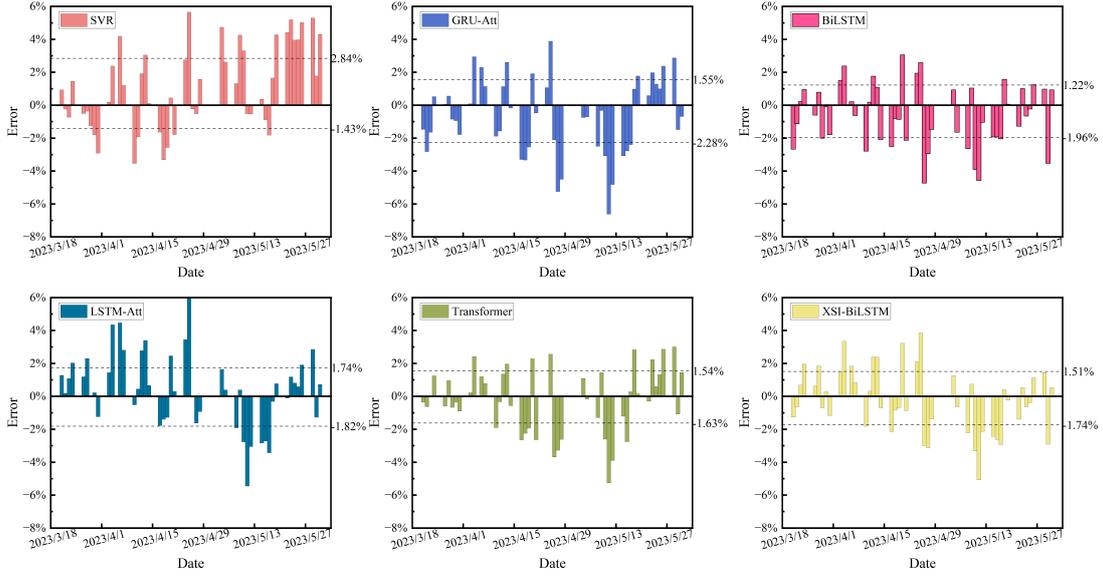

Fig. 9 RPE of each model for SZ600519

Fig. 10 RPE of each model for SZ300750

**5.4 Ablation experiment**

In the prediction module of XSI-BiLSTM, we incorporate an investor sentiment index and introduce a highway mechanism to enhance the information-capturing ability of the BiLSTM model. Therefore, it is necessary to verify its effectiveness through ablation experiments. Table 8 shows the performance comparison results between BiLSTM, BiLSTM using sentiment index (SI), BiLSTM using the highway mechanism, and both (i.e., the modules we use). The results are presented in multiplicative form using BiLSTM performance as a benchmark.

Table 8 Performance comparison results of the ablation experiment

| Stock | Input windows | 7 | | | 15 | | | 30 | | |
|---|---|---|---|---|---|---|---|---|---|---|
| | Model | RMSE | MAPE | R2 | RMSE | MAPE | R2 | RMSE | MAPE | R2 |
| SZ399006 | BiLSTM | 1x | 1x | 1x | 1x | 1x | 1x | 1x | 1x | 1x |
| | BiLSTM-SI | 1.18x | 1.14x | 1.16x | **1.23x** | 1.20x | **1.15x** | 1.22x | 1.14x | 1.26x |
| | BiLSTM-highway | 1.29x | 1.23x | 1.25x | 1.04x | 1.01x | 1.04x | 1.25x | **1.21x** | 1.31x |
| | Our | **1.35x** | **1.37x** | **1.26x** | 1.19x | **1.21x** | 1.12x | **1.29x** | 1.20x | **1.32x** |
| SZ399001 | BiLSTM | 1x | 1x | 1x | 1x | 1x | 1x | 1x | 1x | 1x |
| | BiLSTM-SI | 1.16x | 1.14x | 1.03x | 1.20x | **1.22x** | 1.04x | **1.16x** | **1.15x** | **1.03x** |
| | BiLSTM-highway | 1.17x | 1.19x | **1.04x** | 1.23x | 1.16x | **1.05x** | 1.11x | 1.08x | 1.02x |
| | Our | **1.24x** | **1.23x** | **1.04x** | **1.24x** | 1.19x | 1.04x | 1.13x | 1.11x | 1.02x |
| SH600519 | BiLSTM | 1x | 1x | 1x | 1x | 1x | 1x | 1x | 1x | 1x |
| | BiLSTM-SI | 1.07x | 1.14x | 1.01x | **1.15x** | 1.08x | 1.06x | 1.07x | 1.13x | 1.00x |
| | BiLSTM-highway | 1.09x | 1.12x | **1.07x** | 1.13x | 1.07x | **1.10x** | 1.02x | 1.08x | 1.02x |
| | Our | **1.13x** | **1.21x** | 1.05x | **1.15x** | **1.10x** | 1.07x | **1.14x** | **1.23x** | **1.05x** |
| SZ300750 | BiLSTM | 1x | 1x | 1x | 1x | 1x | 1x | 1x | 1x | 1x |
| | BiLSTM-SI | 1.12x | 1.14x | 1.18x | 1.10x | **1.11x** | 1.17x | 0.98x | 0.99x | 1.02x |
| | BiLSTM-highway | 1.12x | 1.09x | 1.14x | 1.05x | 1.03x | 1.06x | **1.02x** | **1.02x** | 1.03x |
| | Our | **1.16x** | **1.16x** | **1.22x** | **1.11x** | 1.10x | **1.18x** | 1.01x | **1.02x** | **1.06x** |

As shown in Table 8, our module achieves optimal prediction performance in most scenarios, especially when the input sequence length is short (input window = 7). The model's ability to mine the data is enhanced by adding investor sentiment and highway methods, which help extract time-series features. Furthermore, as the input window expands, the degree of performance improvement from both diminishes, showing that long-term time series may compensate for the lack of external information and help the neural network anticipate stock values.

Fig. 11 shows the average percentage improvement of each model compared to BiLSTM for different input windows and metrics. For shorter input windows, the BiLSTM-highway improves better than the BiLSTM-SI; for medium-length input windows, the BiLSTM-SI improves better than the BiLSTM-highway; and for longer input windows, the improvements are similar. These situations may be because the highway mechanism is more effective in dealing with short-term dependencies, as it allows information to be passed more directly between layers, reducing the risk of information loss. Moreover, as the input window length increases, the introduction of investor sentiment can help the model capture long-term market sentiment. In summary, the results for each indicator reveal that the combined use of the two always yields better predictions than their individual use.

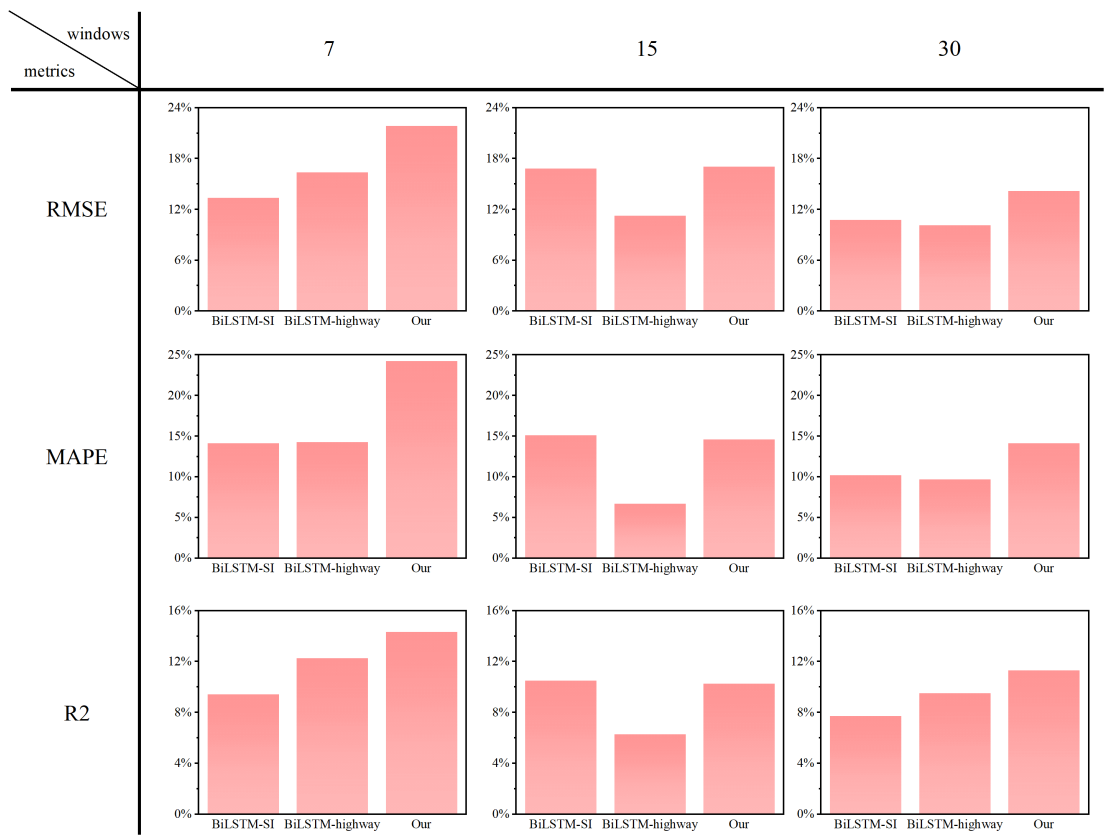

Fig. 11 Performance improvement effect of each model

In addition, Fig. 12 shows a box plot of the absolute error of prediction for each model. The errors always show a skewed distribution, i.e., the difference between the model's predicted price and the actual price is negligible in most cases. The introduction of investor sentiment effectively reduces the median prediction error, while introducing the highway mechanism reduces the extreme values and avoids large errors. After combining the two approaches, the forecast error decreases overall, as reflected by the fact that its average error (i.e., the black square in the figure) is the lowest among all four stocks.

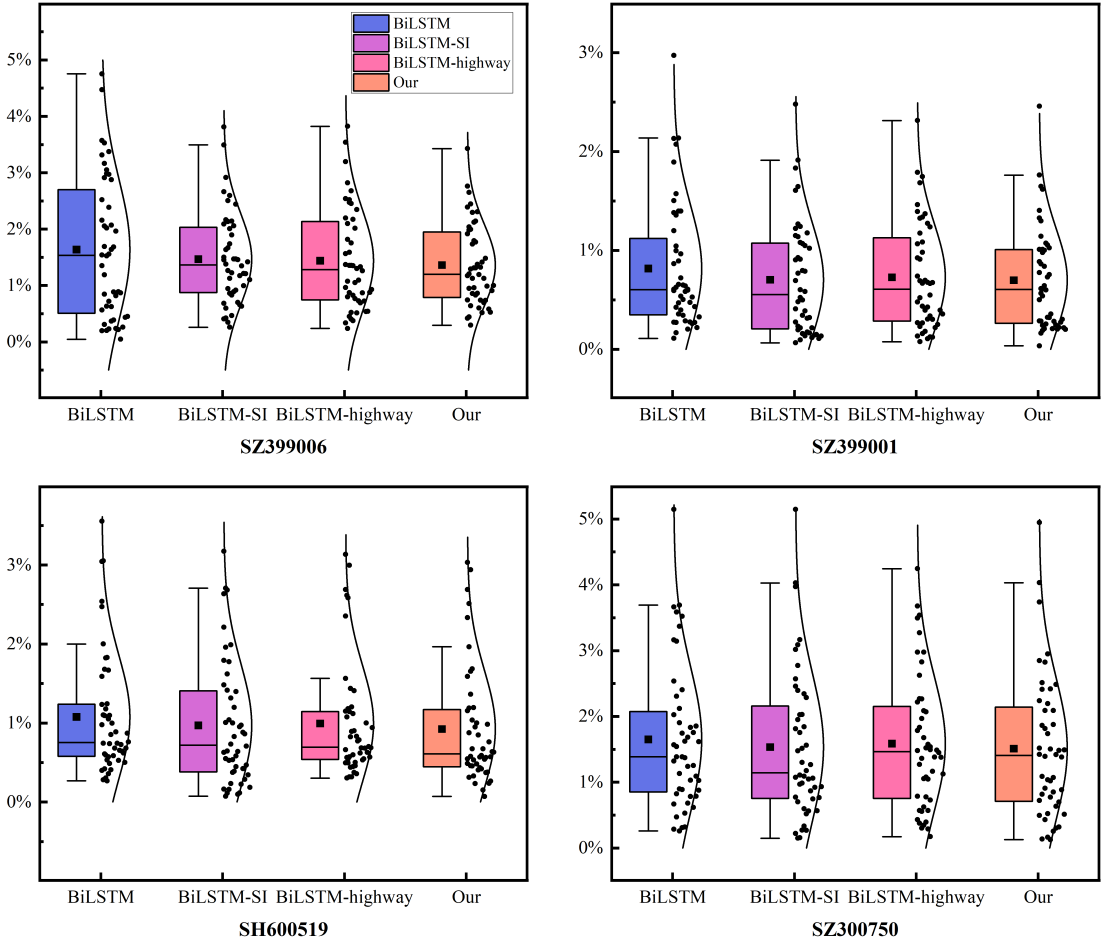

Fig. 12 Box-distribution plots of each model's absolute error

## 6 Conclusion

In this paper, we propose a novel hybrid framework for stock price prediction called XSI-BiLSTM, which improves the computation of daily investor sentiment by analyzing investor sentiment in online forums and taking into account the popularity of posts. Finally, we predict the next day's closing price of a stock using the improved BiLSTM. The effectiveness of XSI-BiLSTM is demonstrated through extensive validation experiments conducted on two constituent indices and two stocks in the Chinese stock market.

Specifically, the experiments show that the XLNET model used by the framework can effectively extract textual sentiment from posts, and the improved sentiment index forms a bidirectional Granger causality with stock price changes to a certain extent, which can better reflect and predict stock price trends. The prediction results of XSI-BiLSTM outperform those of the other prediction models in most of the performance metrics, and the actual stock closing prices differ less from the predicted stock closing prices. In addition, ablation experiments evaluate the effectiveness of the highway mechanism in the improved prediction module and the necessity of introducing investor sentiment.

In this study, we found considerable information noise in stock forums, which does not contribute to stock price prediction. However, views that receive more attention and recognition reflect stock price movements more accurately. Future research will explore in more depth the impact of investor sentiment and its external attributes on stock prices and analyze how to better analyze stock market-related textual information that can help improve model prediction performance.

**CRediT authorship contribution statement**

**Huiyu Li**: Conceptualization, Methodology, Data curation, Software, Writing - original draft. **Junhua Hu**: Funding acquisition, Project administration, Writing – review & editing.

**Data availability**

Data will be available on request.

**Acknowledgments**

This work received support by the Fundamental Research Funds for the Central Universities of Central South University (Grant number: 1053320220429).